\documentclass[12pt, final]{l4dc2023}
\usepackage{times}

\usepackage{bbm, amsmath, amssymb}  %
\usepackage{xcolor}
\usepackage{algorithm}
\usepackage[noend]{algpseudocode}
\usepackage{wrapfig}
\usepackage{comment}

\definecolor{darkgreen}{RGB}{29, 177, 2}
\definecolor{porange}{RGB}{231, 117, 0}  %
\definecolor{nvgreen}{HTML}{76B900}  %

\algrenewcommand\algorithmiccomment[1]{\textcolor{porange}{\hfill$\triangleright$ #1}}
\algnewcommand{\LineComment}[1]{\State \textcolor{porange}{// #1}}

\DeclareMathOperator*{\argmax}{{\mathop{\mathrm{argmax}}}}

\DeclareMathOperator*{\expect}{{\mathbb{E}}}

\newcommand{\sgnDist}[1]{{s_{#1}}}

\newcommand{\reals}{\mathbb{R}}
\newcommand{\naturals}{\mathbb{N}}

\newcommand{\batch}{{\mathcal{B}}}

\newcommand{\state}{{x}}

\newcommand{\statenext}{{\state'}}
\newcommand{\ctrl}{{u}}
\newcommand{\ctrlaux}{{\tilde{\ctrl}}}
\newcommand{\ctrlnext}{{\ctrl'}}

\newcommand{\ctrlimag}[2]{{\ctrl_{#1|#2}}}
\newcommand{\dstb}{{d}}
\newcommand{\dstbaux}{{\tilde{\dstb}}}
\newcommand{\dstbnext}{{\dstb'}}

\newcommand{\traj}{{\mathbf{\state}}}

\newcommand{\stateimag}[2]{{\state_{#1|#2}}}

\newcommand{\xSet}{{\mathcal{X}}} %
\newcommand{\cSet}{{\mathcal{U}}}
\newcommand{\dSet}{{\mathcal{D}}}

\newcommand{\nx}{{n_\state}}
\newcommand{\nc}{{n_\ctrl}}
\newcommand{\nd}{{n_\dstb}}

\newcommand{\dyn}{{f}} %
\newcommand{\dynNom}{{\bar{\dyn}}}

\newcommand{\tdisc}{{t}} %
\newcommand{\tdiscaux}{{\tau}}

\newcommand{\khorizon}{{H}}

\newcommand{\fx}[1]{{\dyn_{\state, #1}}}
\newcommand{\fu}[1]{{\dyn_{\ctrl, #1}}}
\newcommand{\fd}[1]{{\dyn_{\dstb, #1}}}

\newcommand{\deltax}{{\delta\state}}
\newcommand{\deltau}{{\delta\ctrl}}

\newcommand{\outcome}{{J}}  %
\newcommand{\valFunc}{{V}}
\newcommand{\qFunc}{{Q}}

\newcommand{\policy}{{\pi}}

\newcommand{\policySet}{{\Pi}}

\newcommand{\consFunc}{{g}}
\newcommand{\consFuncNext}{{\consFunc'}}

\newcommand{\target}{{\mathcal{T}}}

\newcommand{\failure}{{\mathcal{F}}}
\newcommand{\safeSet}{{\Omega}}
\newcommand{\reach}{{\mathcal{R}}}

\newcommand{\replay}{{\mathcal{B}}}

\newcommand{\policyParam}{{\theta}}
\newcommand{\policyCtrl}{{\policy^\ctrl}}
\newcommand{\policyCtrlAux}{{\tilde{\policy}^\ctrl}}
\newcommand{\policyDstb}{{\policy^\dstb}}
\newcommand{\policyDstbAux}{{\tilde{\policy}^\dstb}}
\newcommand{\policyParamCtrl}{{\policyParam}}
\newcommand{\policyParamDstb}{{\phi}}
\newcommand{\actorCtrl}{{\policy^\ctrl_\policyParamCtrl}}
\newcommand{\actorDstb}{{\policy^\dstb_\policyParamDstb}}
\newcommand{\criticParam}{{\omega}}
\newcommand{\criticauxParam}{{\omega'}}
\newcommand{\criticVal}{{\valFunc_{\criticParam}}}
\newcommand{\critic}{{\qFunc_{\criticParam}}}
\newcommand{\criticaux}{{\qFunc_\criticauxParam}}

\newcommand{\entCoeffCtrl}{{\alpha_\ctrl}}
\newcommand{\entCoeffDstb}{{\alpha_\dstb}}

\newcommand{\softUpdate}{{\lambda_\criticauxParam}}
\newcommand{\updateRatio}{{\tau}}

\newcommand{\task}{{\text{task}}}

\newcommand{\shieldRobust}{{\Delta_{\reach}}}

\newcommand{\shieldNaive}{{\Delta_{\text{ro}}}}
\newcommand{\policyTask}{{\policy^\task}}

\newcommand{\filter}{{\phi}}
\newcommand{\policyShield}{{\filter}}
\newcommand{\policyInv}{{\policy^\target}}

\newcommand{\m}{~\text{m}}

\newcommand{\mss}{~\text{m}/\text{s}^2}

\newboolean{showrevision}
\setboolean{showrevision}{false}
\newcommand{\new}[1]{\ifthenelse{\boolean{showrevision}}{\textcolor{porange}{#1}}{#1}}
\newcommand{\newcaption}[1]{\ifthenelse{\boolean{showrevision}}{\caption{\textcolor{porange}{#1}}}{\caption{#1}}}

\title{ISAACS: Iterative Soft Adversarial Actor-Critic for Safety}
\author{%
 \Name{Kai-Chieh Hsu}\thanks{Equal contribution.} \Email{kaichieh@princeton.edu}\\
 \Name{Duy P. Nguyen}\footnotemark[1] \Email{duyn@princeton.edu}\\
 \Name{Jaime F. Fisac} \Email{jfisac@princeton.edu}\\
 \addr  Department of Electrical and Computer Engineering, Princeton University, NJ, USA
}

\graphicspath{{./figure/}}

\begin{document}

\maketitle

\begin{abstract}
The deployment of robots in uncontrolled environments requires them to operate robustly under previously unseen scenarios, like irregular terrain and wind conditions.
Unfortunately, while rigorous safety frameworks from robust optimal control theory scale poorly to high-dimensional nonlinear dynamics, control policies computed by more tractable ``deep'' methods lack guarantees and tend to exhibit little robustness to uncertain operating conditions.
This work introduces a novel approach enabling scalable synthesis of robust safety-preserving controllers for robotic systems with general nonlinear dynamics subject to bounded modeling error, by combining game-theoretic safety analysis with adversarial reinforcement learning in simulation.
Following a soft actor-critic scheme, a safety-seeking fallback policy is co-trained with an adversarial ``disturbance'' agent that aims to invoke the worst-case realization of model error and training-to-deployment discrepancy allowed by the designer's uncertainty.
While the learned control policy does not intrinsically guarantee safety, 
it is used to construct a real-time safety filter with robust safety guarantees
based on forward reachability rollouts.
This safety filter can be used in conjunction with a safety-agnostic control policy, precluding
any task-driven actions that \emph{could} result in loss of safety.
We evaluate our learning-based safety approach in a 5D race car simulator, compare the learned safety policy to the numerically obtained optimal solution,
and empirically validate the
robust safety guarantee of our proposed safety filter against worst-case model discrepancy.
\end{abstract}


\section{Introduction}
\looseness=-1
Recent years have seen a rapid increase in the deployment of robotic systems
beyond their traditional industrial settings,
with emerging applications including
home robots, autonomous driving, and a range of drone services. %
These new opportunities are tied to \new{open,} uncontrolled %
environments, 
\new{where safe robot operation is at once critical and hard to ensure.}
\new{Safety guarantees in these open-world settings face the coupled challenges of \textit{scalability} and \textit{robustness}.}
\new{Many modern robotic systems present high-order nonlinear dynamics, making safety analysis computationally demanding.}
\new{Even when the analysis is tractable (usually for lower-fidelity models),}
discrepancies between the modeled and \new{physical} system can result in degraded performance and even catastrophic failures.

To ensure 
\new{safe autonomous operation},
a range of engineering efforts seek to
\new{automatically}
\textit{filter}
\new{robots'}
task-oriented \new{control} policies
\new{to preclude} unsafe actions.
\new{One important family of methods is built on}
Hamilton-Jacobi (HJ) reachability analysis,
formulating a zero-sum dynamic game where the \new{robot's} \textit{controller} aims to keep the \new{state} away from \new{all known} failure conditions 
\new{despite the}
adversarial
\new{inputs of a bounded}
\textit{disturbance}
representing
unknown model error \new{and exogenous perturbations}%
~\citep{mitchell2005timedependent}.
While powerful, HJ methods
\new{become computationally prohibitive}
beyond 
5 %
state dimensions~\citep{bansal2017hamilton}.
Recent  %
\new{research with neural representations}
shows promise in scaling safety analysis to high-dimensional systems~\citep{fisac2019bridging, bansal2021deepreach, hsuzen2022sim2lab2real}.
\new{Unfortunately,} the learned 
policy and value function
\new{may not be accurate everywhere,}
\new{and therefore carry a}
risk of catastrophic outcomes if directly 
\new{relied upon for safety.}

\begin{figure}[!t]
\floatconts
    {fig:overview}
    {\vspace{-8mm}}
    {\vspace{-5mm}\includegraphics[width=0.75\textwidth]{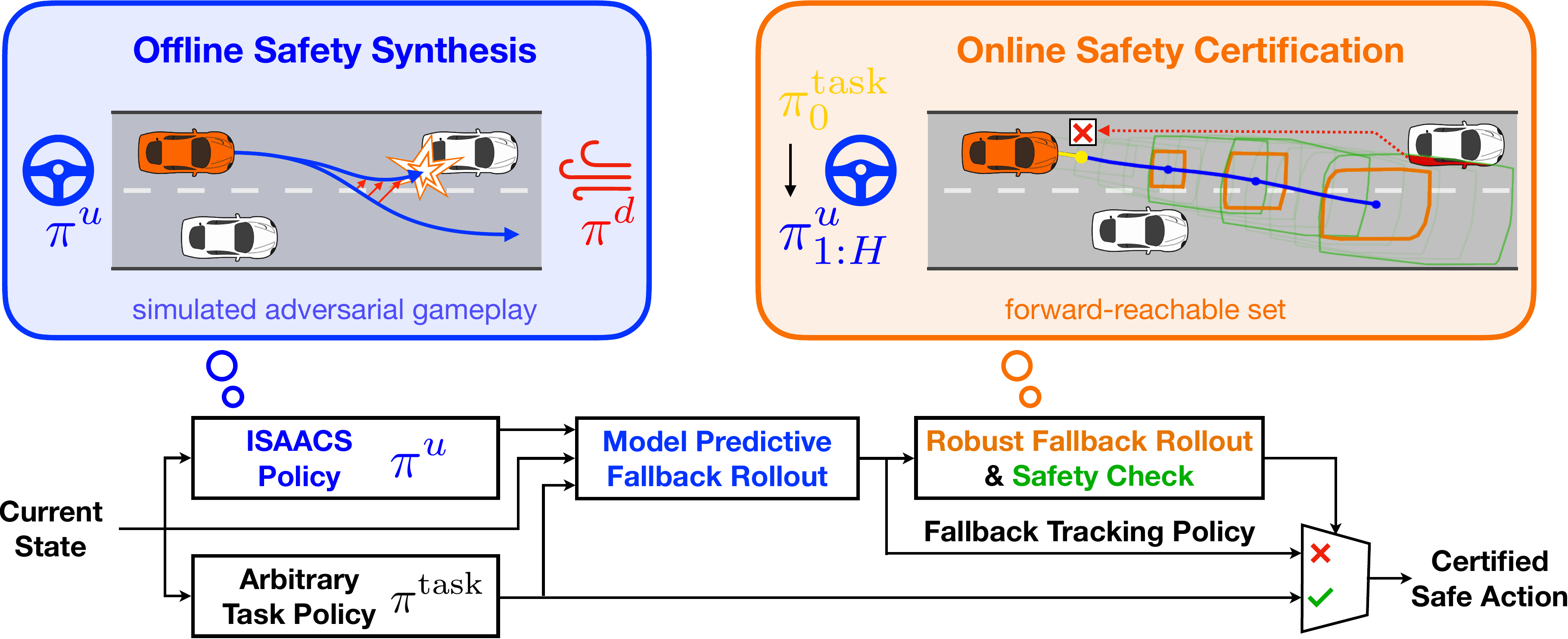}
    }
    \caption{
    ISAACS is a game-theoretic reinforcement learning scheme whose best-effort learned safety policy can be converted into effective robust safety-certified strategies at runtime.
    \textit{Offline Safety Synthesis:} adversarial reinforcement learning approximately solves the robust safety problem, jointly training the safety policy $\policyCtrl$ and worst-case disturbance~$\policyDstb$. \textit{Online Safety Certification:} the learned safety policy is rolled out under all disturbance realizations. Here, the forward-reachable sets (orange) are safe if their footprint-augmented counterparts (green) remain collision-free.
    \textit{Robust Safety Filter:} control actions proposed by an arbitrary task policy $\policyTask$ are allowed if a subsequent safety policy rollout (``fallback'') is certified safe; otherwise, the (already certified) fallback tracking policy is used.
    }
    \vspace{-8mm}
\end{figure}

This \new{paper} introduces Iterative Soft Adversarial Actor-Critic for Safety (ISAACS),
a novel
\new{game-theoretic reinforcement learning (RL) scheme for \emph{approximate safety analysis},
whose outputs can be efficiently converted at runtime into \emph{robust safety-certified control strategies}}
(\figureref{fig:overview}).
\new{ISAACS is first used in an
\emph{offline synthesis} stage \new{that} jointly trains a best-effort safety controller and a worst-case failure-seeking disturbance
through many iterations of simulated zero-sum gameplay~\citep{silver2017alphazero,pinto2017robust}%
.}
\new{The learned control policy can then be treated as an ``untrusted oracle'' and used in \emph{online safety certification} by guiding a robust predictive rollout that accounts for all admissible realizations of model uncertainty.
This ``rollout check'' enables a \emph{recursively safe} runtime control filter that preemptively overrides any candidate
control action that could otherwise drive the state into an unrecoverable configuration.}

\new{We demonstrate this framework on a dynamical system at the boundary of computability for numerical dynamic programming~\citep{bui2022optimizeddp}, namely a 5-dimensional race car simulator.
Rather than substantiating scalability (already established in prior work, cf.~\cite{haarnoja2018sac,fisac2019bridging}), we focus on showing the framework's ability to 
synthesize provably correct robust safety filters using deep reinforcement learning, and use the still-tractable numerical 5-D solution as a reference oracle.
In our experiments, the ISAACS robust rollout safety filter maintained a perfect (zero violation) safety rate under the oracle worst-case disturbance, with moderate conservativeness relative to the optimal Hamilton-Jacobi solution. Using the ISAACS model directly performed well for the most part, but it lacked theoretical guarantees and the safety rate was indeed nonzero.\footnote{See \url{https://saferobotics.princeton.edu/research/isaacs/} for supplementary material.}
}

\subsection{Related Work}
\looseness=-1
Safety guarantees in robotics have their origins in robust control.
Robust ``tube'' model predictive control (MPC) approaches~\citep{langson2004robust} allow enforcing state constraints like collision avoidance in the presence of bounded uncertainty.
Hamilton-Jacobi-Isaacs (HJI) theory handles general nonlinear dynamics and control objectives by posing the problem as a two-player zero-sum differential game between the controller and an adversarial disturbance~\citep{isaacs1954differential,mitchell2005timedependent,bansal2017hamilton}.
Similar approaches pose the game in temporal logic form using 
formal methods%
~\citep{mattila2015iterative,alpern1985defining}.
While these techniques enjoy strong theoretical properties, their use in robotics
is limited by their poor computational scalability.

Control barrier function (CBF) approaches 
aim to circumvent \new{numerical intractability} by finding a smaller (more conservative)
invariant set
that can be encoded in
closed form (e.g., sum-of-squares)~\citep{ames2014control,ames2019control}; unfortunately there are no systematic constructive methods,
so system-specific hand design is often needed~\citep{nguyen2016optimal,squires2018constructive}.
Further, CBF guarantees have limited robustness to disturbances~\citep{xu2015robustness} and may be lost entirely if the robot reaches its actuation limits~\citep{zeng2021safetycritical,choi2021robust}.

Finally, deep self-supervised learning ~\citep{bansal2021deepreach} and reinforcement learning~\citep{fisac2019bridging,bharadhwaj2021conservative,thananjeyan2021recovery,hsu2021safety} can synthesize \new{approximate}
control policies and value functions (``safety critics''),
but offer no \new{intrinsic} safety guarantees, and robustness to modeling and learning error is not yet well understood.
Domain randomization (DR)
\new{approaches~\citep{tobin2017domain, mehta2020active} conduct training under a family of environments with randomly sampled parameter values, targeting expected performance over the hypothesized parameter distribution, whereas}
adversarial training \new{approaches} formulate a zero-sum game and \new{jointly} train
the control policy and the worst-case environment realization
via simulated gameplay~\citep{pinto2017robust}.
While most closely aligned with~\cite{fisac2019bridging,hsu2021safety}, our work introduces robustness through an adversarial reinforcement learning scheme grounded in the game-theoretic HJI formulation and further recovers robust guarantees by rolling out the learned policies in an online \new{receding horizon} framework.

\section{Preliminaries}
\subsection{Hamilton-Jacobi-Isaacs Reachability Analysis and Safety Filters}
We consider fully observable robotic system governed by discrete-time dynamics with unknown but bounded model error:
\begin{equation}
    \state_{\tdisc+1} = \dyn(\state_\tdisc, \ctrl_\tdisc, \dstb_\tdisc)\,, \label{eq:dyn}
\end{equation}
where, at each time step $\tdisc\in\naturals$, $\state_\tdisc \in \xSet \subseteq \reals^\nx$ is the state, $\ctrl_\tdisc \in \cSet \subset \reals^\nc$ is the \textit{controller} input, and $\dstb_\tdisc \in \dSet \subset \reals^\nd$ is the \textit{disturbance} input, unknown \textit{a priori}.
We assume we are given a specification of the \textit{failure set} $\failure \subseteq \xSet$ that the system must be prevented from entering.
By convention, we assume $\failure$ to be open.
Safety analysis seeks to determine
the \textit{safe set} $\safeSet \subseteq \xSet$, consisting of all initial states from which
there exists a control policy that can keep the system away from the failure set at all times for \emph{any} realization of the uncertainty:
\begin{equation}
    \safeSet := \Big\{
        \state \in \xSet \mid \exists \policyCtrl: \xSet \to \cSet,~ \forall \policyDstb: \xSet \to \dSet,~ \forall \tdisc > 0,\; \traj_{\state}^{\policyCtrl\!,\policyDstb}(\tdisc) \notin \failure
    \Big\}\,, \label{eq:safe_set}
\end{equation}\\[-10pt]
where $\traj_{\state}^{\policyCtrl\!,\policyDstb} \colon \naturals \to \xSet$ denotes the trajectory starting from state $\state_0 = \state$ following dynamics~\eqref{eq:dyn} under control policy $\policyCtrl$ and disturbance policy $\policyDstb$. Note that the order of quantifiers in~\eqref{eq:safe_set} is crucial: there must be one (same) control policy $\policyCtrl$ that maintains safety under all disturbance policies $\policyDstb$.

Hamilton-Jacobi-Isaacs (HJI) reachability analysis formulates the robust safety problem as a zero-sum game between the controller and the disturbance, introducing a Lipschitz continuous \textit{safety margin} $\consFunc \colon \xSet \to \reals, \consFunc(\state) < 0 \Longleftrightarrow \state \in \failure$
to further transform the safety ``game of kind'' with a binary outcome (whether the system enters $\failure$) into a ``game of degree'' with continuous payoff%
\begin{equation}
    \outcome^{\policyCtrl\!,\policyDstb}(\state) := \min_{\tdisc \in \naturals} \consFunc \Big(
        \traj_{\state}^{\policyCtrl\!,\policyDstb}(\tdisc)
    \Big)
    \,.
    \label{eq:hj_outcome}
\end{equation}
Consistent with the order of quantifiers in~\eqref{eq:safe_set}, which gives the disturbance the instantaneous informational advantage~\citep{isaacs1954differential},
we define the lower value function of the safety game as $\valFunc(\state) := \max_{\policyCtrl} \min_{\policyDstb} \outcome^{\policyCtrl\!,\policyDstb}(\state)$,
which encodes the minimal safety margin $\consFunc$ that our controller can maintain at all times under the worst-case disturbance.
This value $\valFunc(\state)$ satisfies the two-player dynamic programming Isaacs equation
\begin{equation}
    \valFunc(\state) = 
    \max_{\vphantom{\dstb}\ctrl} \min_{\dstb}
    \min\Big\{\consFunc(\state), \, 
    \valFunc\big( \dyn(\state,\ctrl,\dstb)   \big) \Big\}
    \,.
    \label{eq:isaacs}
\end{equation}
If the value function can be computed, the safe set~\eqref{eq:safe_set} can be obtained by $\valFunc(\state) \geq 0 \Longleftrightarrow \state \in~\safeSet$,
and the optimal policies $\policy^{\ctrl*}(\state)$, $\policy^{\dstb*}(\state)$ are given by the optimizers of~\eqref{eq:isaacs} at each state~$\state$.
We can \new{then} \textit{filter}
an arbitrary task-oriented policy $\policyTask$
through the least-restrictive law \citep{fisac2019AGS}:
\begin{equation}
    \policyShield(\state; 
    \policyTask
    ) =
    \begin{cases}
            \policyTask (\state), & 
            \valFunc(\state) \geq \epsilon \\
            \policy^{\ctrl*} (\state), & \text{otherwise}
    \end{cases}
    \label{eq:shield_original}
\end{equation}
where $\epsilon \ge 0$ is the value threshold, typically chosen slightly larger than zero to account for numerical errors and control delays.
The safety filter $\policyShield$ enforces an important invariance property: from any state in the safe set $\safeSet$ the system is guaranteed to remain in the safe set \new{perpetually}. %

\subsection{Reachability Analysis through Reinforcement Learning}
Level-set methods solve for the HJI value function in \eqref{eq:isaacs} with vanishing approximation error as the grid resolution increases.
However, the memory and computation complexity grows exponentially with the state dimension, which limits \new{practical} applicability to dynamical systems with at most 6 continuous state dimensions \citep{bui2022optimizeddp}.
\citet{fisac2019bridging} use reinforcement learning algorithms to more tractably find approximate solutions to high-dimensional reachability problems
(demonstrated on up to 18 state dimensions)
by
replacing the usual reinforcement learning Bellman equation for a cumulative reward with the time-discounted (single-player) counterpart of~\eqref{eq:isaacs}:
\new{
\begin{equation}
   \criticVal(\state) =
        (1-\gamma) \consFunc(\state) + 
        \gamma \min \Big\{
            \consFunc(\state),~
            \max_{\ctrl \in \cSet} \critic (\state, \ctrl)
        \Big\}, \quad
    \critic (\state, \ctrl) := \criticVal \left( \dynNom(\state, \ctrl) \right)\,,
    \label{eq:hjb_dis_value}
\end{equation}\\[-10pt]
}%
where \new{the nominal dynamics $\dynNom \colon \xSet \times \cSet \to \xSet$, which can be seen as a special case of $\dyn$ in~\eqref{eq:dyn} assuming no modeling error ($\dstb=0$); $\gamma \in (0, 1)$ is the time discount rate for future safety margins; and $\critic$ is the state-action safety value function, parameterized by $\criticParam$.}\footnote{In the deep reinforcement learning literature,  $\criticParam$ are neural network weights and $\critic$ is called a Q-network or critic.}
An analogous safety filter to~\eqref{eq:shield_original} can then be constructed by replacing $\valFunc$ with the learned $\criticVal$
and $\policy^{\ctrl*}(\state)$ with $\argmax_{\ctrl \in \cSet} \critic(\state, \ctrl) $.
\new{Unfortunately, due to the approximate nature of $\critic$ and $\criticVal$, this is only a \emph{best-effort} safety filter:} unlike~\eqref{eq:shield_original}, it comes with no invariance guarantees
\new{and it cannot generally prevent safety violations}.

\section{ISAACS: Iterative Soft Adversarial Actor-Critic for Safety}
\new{To harness the scalability of neural representations
without renouncing the robust safety guarantees of model-based analysis,}
we propose Iterative Soft Adversarial Actor-Critic for Safety (ISAACS),
a game-theoretic reinforcement learning scheme that approximates the HJI solution to a reachability game %
and learns a safety policy that can be used to construct a provably safe runtime control strategy. 
ISAACS uses repeated rounds of simulated zero-sum gameplay to
jointly train a safety control policy and a failure-seeking disturbance policy, consistent with the \new{Isaacs equation}~\eqref{eq:isaacs}.
Once trained, the ISAACS safety policy can be used as the receding-horizon reference for a robust fallback control strategy, defining a compact forward-reachable set that can be checked for safety violations.
This results in a recursive safety filter with an equivalent invariance property to the accurate but less scalable counterpart~\eqref{eq:shield_original} enabled by numerical HJI methods.

\subsection{Adversarial Actor-Critic Reinforcement Learning for Safety Policy Synthesis}
\begin{algorithm}[!b]
\footnotesize
\caption{\small ISAACS: Iterative Soft Adversarial Actor-Critic for Safety (Offline Safety Synthesis)} \label{alg:ISAACS}
\begin{algorithmic}[1]
\For{each tournament round}
    \For{each episode}
        \State $\state_0 \sim P_0, \policyDstbAux \sim P_{\policySet^\dstb}$ \Comment{Sample initial state and disturbance policy for this episode}
        \For {each time step}
            \State $\ctrl_\tdisc \sim \actorCtrl (\cdot | \state_\tdisc)$, $\dstb_\tdisc \sim \policyDstbAux (\cdot | \state_\tdisc)$ \Comment{Sample control and disturbance from policies}
            \State $\state_{\tdisc+1}=\dyn(\state_\tdisc, \ctrl_\tdisc, \dstb_\tdisc)$ \Comment{Get transition from the environment}
            \State $\replay \gets \replay \cup \big\{\big(\state_\tdisc, \ctrl_\tdisc, \dstb_\tdisc, \state_{\tdisc+1}, \consFunc(\state_{\tdisc+1})\big)\big\}$ \Comment{Store the transition in the replay buffer}
        \EndFor
        \For {each gradient step}
            \State $\criticParam \gets \criticParam - \lambda_\criticParam \nabla_\criticParam L(\criticParam)$ \Comment{Update critic parameters}
            \State $\criticauxParam \gets \softUpdate \criticParam + (1-\softUpdate)\criticauxParam$ \Comment{Update target critic parameters}
            \State $\policyParamDstb \gets \policyParamDstb-\lambda_\policyParamDstb \nabla_\policyParamDstb L(\policyParamDstb)$ \Comment{Update disturbance policy parameters}
            \If {gradient step is a multiple of $\updateRatio$}
                \State $\policyParamCtrl \gets \policyParamCtrl - \lambda_\policyParamCtrl \nabla_\policyParamCtrl L(\policyParamCtrl)$ \Comment{Update safety policy parameters at a slower rate}
            \EndIf
        \EndFor
    \EndFor
    \State Update leaderboard $\policySet^\ctrl \bigcup \{\actorCtrl\}$ vs. $\policySet^\dstb \bigcup \{\actorDstb\}$ and keep best $k^\ctrl$ in $\policySet^\ctrl$, best $k^\dstb$ in $\policySet^\dstb$
    \State $m \gets (m_1, \cdots, m_{|\policySet^\dstb|})$ \Comment{$m_i$ is the total win rate of $\policy^\dstb_i$ across all $\policyCtrlAux\in\policySet^\ctrl$}
    \State $P_{\policySet^\dstb} \gets \text{softmax}(m)$ \Comment{Update disturbance policy distribution}
\EndFor 
\end{algorithmic}
\end{algorithm}
Analogous to the single-player reachability reinforcement learning formulation of \cite{fisac2019bridging},
we consider a time-discounted counterpart of the reachability payoff~\eqref{eq:hj_outcome}.
In this case, however, we have a zero-sum game whose value function is
characterized by a two-player Isaacs equation (rather than a Bellman equation).
In the space of soft actor-critic policies,
the Isaacs equation can be written
\new{
\begin{equation}
    \valFunc%
    (\state) =
        (1-\gamma) \consFunc(\state) + \gamma 
        \max_{\vphantom{\policyDstb} \policyCtrl} \min_{\policyDstb}
        \expect_{\vphantom{\policyDstb} \ctrl, \dstb}
        \min\Big\{
            \consFunc(\state),
            \qFunc (\state, \ctrl, \dstb)
        \Big\},
    \quad
    \qFunc (\state, \ctrl, \dstb) :=
        \valFunc
            \big(
                \dyn (\state, \ctrl, \dstb)
            \big)
    \label{eq:hji_dis_value}
\end{equation}\\[-10pt]
\noindent%
where $\policyCtrl,\policyDstb$ are \emph{stochastic} controller and disturbance policies, and $\ctrl \sim \policyCtrl(\cdot \mid \state),\dstb \sim \policyDstb(\cdot \mid \state)$.
Note that as the time discount factor $\gamma\in[0,1)$ goes to 1, we recover the undiscounted problem~\eqref{eq:isaacs}.
}

The ISAACS offline synthesis scheme solves the Isaacs equation~\eqref{eq:hji_dis_value} approximately by training three neural networks,
with parameters $\criticParam, \policyParamCtrl, \policyParamDstb$, that encode a critic, a control policy, and a disturbance policy, respectively.
The learning scheme updates the critic and disturbance policy $\tau\in\naturals$ more often than the control policy.
This effectively makes the disturbance policy a \textit{follower} to the control policy~\citep{tijana2021who}---thereby maintaining the disturbance's informational advantage from~\eqref{eq:safe_set} and \eqref{eq:hji_dis_value}---and has the advantage of optimizing against a static target within each update epoch of the controller's policy.

\new{We additionally maintain a finite \emph{leaderboard} of controller and disturbance policies from past stages of training, $\policySet^j = \{\pi^j_1,...\pi^j_{k^j}\}$, $j\in\{u,d\}$.}
At the start of each episode, ISAACS \new{samples an initial state from a preset distribution~$P_0$ and} selects a disturbance policy $\policyDstbAux$ from $\policySet^\dstb$
and \new{simulates the gameplay} between $\actorCtrl$ and the sampled $\policyDstbAux$, which \new{discourages} the control policy updates from overfitting to a single disturbance policy~\citep{vinitsky2020robust}.
Periodically during training, the leaderboard is updated by incorporating the current controller and disturbance policies into $\policySet^\ctrl, \policySet^\dstb$ and simulating multiple gameplay episodes for each new pair of controller-disturbance policies, recording the fraction of episodes that result in safety failures.
Policies in $\policySet^\ctrl$ are ranked based on their overall win rate against all opponent policies in $\policySet^\dstb$, and vice versa, and the worst-performing one is dropped from each leaderboard (if the total count exceeds the preset capacity $k^\ctrl$,$k^\dstb$).

We update all neural networks based on Soft Actor-Critic (SAC)~\citep{haarnoja2018sac}. %
\new{
At every time step we store the transition $(\state, \ctrl, \dstb, \statenext, \consFuncNext)$ in the replay buffer $\batch$, with ${\statenext=\dyn(\state,\ctrl,\dstb)}$ and $\consFuncNext=\consFunc(\statenext)$.
\begin{subequations}
We update the critic to reduce the deviation from the Isaacs target%
~\eqref{eq:hji_dis_value},\footnote{Deep RL usually trains an auxiliary target critic $\criticaux$, whose parameters $\criticauxParam$ change slowly to match the critic parameters $\criticParam$ to stabilize the regression (a fixed target in a short period of time).}
    \begin{equation}
        L(\criticParam) := \expect_{(\state, \ctrl, \dstb, \statenext, \consFuncNext%
        ) \sim \batch} \left[
            \left( \critic(\state, \ctrl, \dstb) - y \right)^2
        \right]\,,
        \quad
        y = (1-\gamma) \consFuncNext + \gamma \min \{ \consFuncNext,~ \criticaux ( \statenext, \ctrlnext, \dstbnext ) \}
        \,,
    \end{equation} 
with $\ctrlnext \sim \actorCtrl(\cdot \mid \statenext)$, $\dstbnext \sim \actorDstb(\cdot \mid \statenext)$.
We update both policies following the policy gradient \new{induced} by the critic and entropy \new{loss terms}: %
    \begin{equation}
        L(\policyParamCtrl) := \!\!\!\! \expect_{(\state, \dstb) \sim \batch} \Big[
            -\critic(\state, \ctrlaux, \dstb) + \entCoeffCtrl \log \actorCtrl(\ctrlaux | \state)
        \Big],
        \;\;
        L(\policyParamDstb) := \!\!\!\! \expect_{(\state, \ctrl) \sim \batch} \Big[
            \critic(\state, \ctrl, \dstbaux) + \entCoeffDstb \log \actorDstb(\dstbaux | \state)
        \Big],
    \end{equation}
where $\ctrlaux \sim \actorCtrl(\cdot \mid \state)$, $\dstbaux \sim \actorDstb(\cdot \mid \state)$, and $\entCoeffCtrl, \entCoeffDstb$ are hyperparameters encouraging higher entropy in the stochastic policies (more exploration), which decay gradually in magnitude through the ISAACS training.
\label{eq:loss}
\end{subequations}
We summarize the ISAACS scheme in \algorithmref{alg:ISAACS} (where $\lambda_\criticParam, \softUpdate, \lambda_\policyParamCtrl, \lambda_\policyParamDstb$ are the learning rate hyperparameters for $\criticParam, \criticauxParam, \policyParamCtrl, \policyParamDstb$).
}

\subsection{Runtime Safety Filter through Robust Policy Rollout}
\new{
It may seem tempting to use the value and policies computed by ISAACS directly in an online safety solution.
While a learned value-based safety filter in the form of~\eqref{eq:shield_original}
can work well in practice~\citep{hsuzen2022sim2lab2real},
it comes without guarantees, and it may not always prevent catastrophic failures.
A similar issue arises when directly rolling out the learned controller and disturbance policies at runtime and checking the resulting trajectory for future collisions:
while a suboptimal controller policy would merely result in a more conservative filter,
a suboptimal disturbance policy may lead us to erroneously conclude that a state is safe, when in reality there exists a different uncertainty realization that could drive the state to the failure set.
To obtain a robust safety guarantee under possibly suboptimal ISAACS learning,
we therefore treat the controller and the disturbance differently.
The learned controller policy is used as a best-effort ``untrusted oracle'' to obtain a reference rollout trajectory;
conversely, for the disturbance, we consider \textit{all} possible inputs within the bounded set~$\dSet$, which induce a forward-reachable set (FRS) containing a continuum of possible futures.
}

\new{At each control cycle, given a proposed control from the task policy, we start by rolling out a nominal reference trajectory with
zero disturbance input.
Similar to other model predictive safety filtering approaches~\cite{bastani2021safe},
we simulate a single step using the proposed control 
and subsequently switch to
the trained safety policy for the remaining $\khorizon$ steps:} 
\begin{align}
    \stateimag{\tdiscaux+1}{\tdisc} &= \dyn(\stateimag{\tdiscaux}{\tdisc}, \ctrlimag{\tdiscaux}{\tdisc}, 0),
    \qquad
    \ctrlimag{\tdiscaux}{\tdisc} =
        \begin{cases}
            \policyTask \left( \stateimag{\tdiscaux}{\tdisc} \right), & \tdiscaux = 0 \\
            \actorCtrl \left( \stateimag{\tdiscaux}{\tdisc} \right), & \tdiscaux \in \{1,\dots,\khorizon\},
        \end{cases}
    &
    \stateimag{0}{\tdisc} &= \state_\tdisc,
    \label{eq:imag_rollout}
\end{align}
\new{where 
$(\cdot)_{\tdiscaux | \tdisc}$ denotes variables at step $\tdiscaux$ of a plan computed at time $\tdisc$.
The blue polyline in \figureref{fig:overview} (\textit{right}) shows the nominal trajectory.
Since the neural network safety policy is not guaranteed to be stabilizing}, we utilize the time-varying linear quadratic regulator (LQR) approach to derive (local) linear tracking policies for the time horizon~$\khorizon$.

To compute the tracking policy, we first linearize the dynamics around the nominal trajectory 
$\deltax_{\tdiscaux+1} =
    \fx{\tdiscaux} \deltax_\tdiscaux +
    \fu{\tdiscaux} \deltau_\tdiscaux +
    \fd{\tdiscaux} \dstb_{\tdisc+\tdiscaux} +
    e_\tdiscaux
$,
at each $\tdiscaux \in \{0,\dots,\khorizon\}$,
where $ {\deltax_{\tdiscaux} := \state_{\tdisc+\tdiscaux} - \stateimag{\tdiscaux}{\tdisc}},$
$\deltau_{\tdiscaux} :=  \ctrl_{\tdisc+\tdiscaux} -  \ctrlimag{\tdiscaux}{\tdisc},
$
and $\fx{\tdiscaux}, \fu{\tdiscaux}, \fd{\tdiscaux}$ are the Jacobians of the dynamics evaluated at prediction step $\tdiscaux$,
with $e_\tdiscaux$ denoting the associated linear approximation error (``Taylor remainder'') at that time.
Using time-varying LQR, we compute the \emph{fallback tracking policy} $\deltau_\tdiscaux = K_{\tdiscaux | \tdisc} \deltax_\tdiscaux, {\tdiscaux \in \{1,\dots,\khorizon\}}$,
which aims to efficiently track the nominal rollout trajectory from~\eqref{eq:imag_rollout}.
The closed-loop linear error dynamics under this policy (with the unknown disturbance as the only exogenous input) are
\begin{equation}
    \deltax_{\tdiscaux+1} = A_\tdiscaux \deltax_\tdiscaux + B_\tdiscaux \dstb_{\tdisc+\tdiscaux} + e_\tdiscaux,~ \tdiscaux \in \{1,\dots,\khorizon\}, \label{eq:linear_dstb_dyn}
\end{equation}
where $A_\tdiscaux := \fx{\tdiscaux} + \fu{\tdiscaux} K_{\tdiscaux | \tdisc}$ and $B_\tdiscaux := \fd{\tdiscaux}$.
Similarly letting $B_0 := \fd{0}$, the forward-reachable set containing all possible (under $\dSet$) tracking errors $\deltax_\tdiscaux$ at each step~$\tdiscaux$ can be computed by
\begin{align}
    \reach_{\tdiscaux+1} & = A_\tdiscaux \reach_\tdiscaux \oplus B_\tdiscaux \dSet
    \oplus \mathcal{E}_\tdiscaux,
    \quad
    \tdiscaux \in \{1,\dots,\khorizon\},
    &
    \reach_{1} & = B_0 \dSet,
    \label{eq:frs}
\end{align}
\new{
where 
$\mathcal{E}_\tdiscaux$ is the bounding box for the Taylor remainder at time step $\tdiscaux$, and $\oplus$ denotes the Minkowski sum of two sets.
The orange polytopes in \figureref{fig:overview} (right) show the computed forward-reachable sets, and the green polytopes show their footprint-augmented counterparts.}

Using the tracking error bounds~$\reach_\tdiscaux$, we define the robust rollout-based safety filter criterion: %
\begin{equation}
    \shieldRobust(\state_\tdisc, \policyTask, \khorizon) := 
    \mathbbm{1}\left\{ 
    \{\stateimag{\tdiscaux+1}{\tdisc}\}
    \oplus
    \reach_{\tdiscaux+1}
    \cap \failure = \emptyset 
    \land
    \{\ctrlimag{\tdiscaux}{\tdisc}\}
    \oplus
    K_\tdiscaux \reach_\tdiscaux
    \subseteq \cSet, 
    \forall \tdiscaux \in \{0, \cdots, \khorizon\} \right\}. \label{eq:shield_robust}
\end{equation}
Precisely, $\shieldRobust(\state_\tdisc, \policyTask, \khorizon) = 1$ means that after applying the proposed control from the task policy,
\new{the tracking policy $(K_{\tdiscaux | \tdisc})_{\tdiscaux=1}^{H}$
}
can maintain safety for the $\khorizon$ subsequent steps under all possible uncertainty realizations, i.e. for any disturbance sequence satisfying $\dstb_{\tdisc+\tdiscaux} \in \dSet$, \new{without exceeding the control bound $\cSet$}.%
\footnote{In practice, the control condition in~\eqref{eq:shield_robust} can be enforced during the rollout of $\reach_\tau$ by scaling down $K_{\tdiscaux | \tdisc}$ as needed.}
In other words, a \emph{robust safety fallback strategy} is available after applying the proposed control.
The corresponding safety filter can be constructed as follows:
\begin{equation}
    \policyShield(\state_{\tdisc+\tdiscaux}; \shieldRobust, \tdisc) = 
        \new{
        \begin{cases}
            \policyTask(\state_{\tdisc+\tdiscaux}), & \shieldRobust(\state_{\tdisc+\tdiscaux}, \policyTask, \khorizon) = 1\\
            K_{\tdiscaux | \tdisc} (\state_{\tdisc+\tdiscaux} - \stateimag{\tdiscaux}{\tdisc}), & \shieldRobust(\state_{\tdisc+\tdiscaux}, \policyTask, \khorizon) = 0 \land \tdiscaux \in \{1, \cdots, \khorizon\} \\
            \actorCtrl(\state_{\tdisc+\tdiscaux}), & \text{otherwise,}
        \end{cases}
        }
        \label{eq:shield_policy}
\end{equation}
where $\tdisc$ is the last time step that the safety filter criterion holds, i.e., $\shieldRobust(\state_\tdisc, \policyTask, \khorizon) = 1$.
We then have the following finite-horizon safety theorem, which is recursively enforceable by applying the safety filter with robust rollout-based criterion in \eqref{eq:shield_robust}.
\new{
\begin{theorem}[Finite-Horizon Safety]
    \small
    If the initial state $\state_\tdisc$ satisfies $\shieldRobust(\state_\tdisc, \actorCtrl, \khorizon)=1$, the safety filter $\policyShield(\cdot; \shieldRobust, \tdisc)$ in \eqref{eq:shield_policy} keeps the feedback system safe under the disturbance set $\dSet$ for at least $\khorizon+1$ steps, i.e., $\state_{\tdisc+\tdiscaux+1} = \dyn(\state_{\tdisc+\tdiscaux}, \policyShield(\state_{\tdisc+\tdiscaux}; \shieldRobust, \tdisc), \dstb_{\tdisc+\tdiscaux}) \notin \failure, \forall \dstb_{\tdisc+\tdiscaux} \in \dSet, \forall \tdiscaux \in \{0, \cdots, \khorizon\}$ .
    \label{thm:finite_hor_safety}
\end{theorem}
\begin{proof}
    \small
    Whenever $\shieldRobust(\state_{\tdisc}, \policyTask, \khorizon) = 1$, a robust $H$-step fallback tracking policy is constructed and found to keep $\{\state_{\tdiscaux|t}\} \oplus \reach_\tdiscaux$ disjoint from $\failure$ for $\tdiscaux\in\{1,\dots,H+1\}$. By construction, after applying $\policyTask(\state_{\tdisc})$, $\forall \dstb_\tdisc\in\dSet, \state_{\tdisc+1}\in\{\state_{1|t}\} \oplus \reach_1$.
    Suppose that all subsequent checks fail: $\shieldRobust(\state_{\tdisc+\tdiscaux}, \policyTask, \khorizon) = 0, \forall \tdiscaux\ge 1$, then the filter~\eqref{eq:shield_policy} applies the fallback policy, ensuring 
    $\forall \dstb_{\tdisc+\tdiscaux}\in\dSet, \state_{\tdisc+\tdiscaux+1}\in\{\state_{\tdiscaux+1|\tdisc}\} \oplus \reach_{\tdiscaux+1}$ up until the last computed set $\reach_{\khorizon+1}$. 
    Therefore, $\state_{\tdisc+1},\dots,\state_{\tdisc+\khorizon+1}\not\in\failure$.
    If at any time $\tdiscaux\le\tdisc+\khorizon+1$, $\shieldRobust(\state_{\tdiscaux}, \policyTask, \khorizon) = 1$, a new fallback tracking policy is computed and the guarantee resets for another $\khorizon+1$ steps.
\end{proof}
}

\vspace{-3mm}
\begin{remark}
If we further assume a robust controlled-invariant set $\target \subseteq \xSet$, $\target \cap \failure = \emptyset$ under policy $\policyInv$, the guarantee can be extended to the infinite horizon, by adding an additional condition that the system robustly reach $\target$ within the rollout horizon, i.e., 
$\reach_\tdiscaux \oplus \{\stateimag{\tdiscaux}{\tdisc}\} \subseteq \target$ for some $\tau\le H+1$.
After this, the controller can always continue to apply $\pi_\target$
to remain in $\target$ and thus out of $\failure$.
Alternatively,
more sophisticated robust certification methods~\citep{wabersich2018linear}
may be used.
\end{remark}
\vspace{-4mm}
\section{Experimental Evaluation}

\subsection{Implementation Details}
\newcommand{\dstbBound}{{\dstb_{\text{max}}}}
\paragraph{Environment}
We demonstrate our framework in a straight-road environment. We consider the uncertain dynamics of a \new{small robot} car modified from a 5D kinematic bicycle model as
\begin{equation}
    \dot \state = [\dot p_x, \dot p_y, \dot v, \dot \psi, \dot \delta] = \left[
        v \cos \psi + \dstb_x, v \sin \psi + \dstb_y, a + \dstb_v, \frac{v}{L} \tan \delta + \dstb_\psi, \omega+\dstb_\delta
    \right], \label{eq:cont_bic_dyn}
\end{equation}
where $(p_x, p_y)$ is the \new{car's} position, $v$ is the forward speed, $\psi$ is the heading, $\delta$ is the steering angle, $L=0.5 \m$ is the wheelbase, $a \in [-3.5, 3.5] \mss$ is the acceleration control, $\omega \in [-5, 5] \text{ rad/s}$ is the steering angular velocity control, $\dstb \in \dSet := \{ \dstbaux \in \reals^5 : \|\dstbaux\|_\infty \le \dstbBound\}$, and $\dstbBound$ is the disturbance bound.
The discrete-time dynamics from \eqref{eq:cont_bic_dyn} are computed by fourth-order Runge-Kutta (RK4) with time step $0.1$~s and implemented in JAX~\citep{bradbury2018jax}. The footprint of the car is represented by a box
$B = [0., 0.5] \times [-0.1, 0.1] \m$, rotated by the car's heading angle and translated by its position: $B(\state) := R_\psi B \oplus \{(p_x,  p_y)\}$.
We consider three constraints: heading angle, road boundary, and obstacles. The safety margin function is defined as\\[-16pt]
\begin{align*}
    \consFunc(\state) & = \min \left\{
        \consFunc_\psi(\state), \consFunc_{\text{road}}(\state), \consFunc_{\text{obs}}(\state)
    \right \},&
    \consFunc_\psi(\state) &= \textstyle\frac{\pi}{2} - |\psi|,
    \\
    \consFunc_{\text{road}}(\state) & = \min_{p \in B(x)} 0.6 - |p_y|,&
    \consFunc_{\text{obs}}(\state) & = \min_{i \in [5]} \min_{p \in B(\state)} \sgnDist{B^i} (p),
\end{align*}\\[-10pt]
where 
$\sgnDist{\mathcal{P}} \colon \reals^2 \to \reals$ is the signed distance function to a nonempty set $\mathcal{P}$ and $B^i := B \oplus \{p^i\}$ are box obstacles at different locations.
A bird's-eye view of the environment can be seen in \figureref{fig:isaacs}.

\paragraph{ISAACS and Baselines}
We initialize the control policy of ISAACS and SAC-DR by training a standard SAC in the absence of a disturbance. Then, we initialize ISAACS' disturbance policy by training another SAC to attack the fixed initial control policy.
The length of the rollout episode is 200 steps (20 seconds) for all RL algorithms.
All policy networks have three fully-connected (FC) layers with 256 neurons, and the critic networks have three FC layers with 128 neurons.
\new{This amounts to 69,634 parameters or 284 KB of storage.}
In SAC-DR we sample the disturbance uniformly from the set~$\dSet$, while in ISAACS we sample it from the (stochastic) disturbance policy.
\new{Finally},
we use resolution-complete finite-difference methods~\citep{bui2022optimizeddp} to solve the HJI equation~\eqref{eq:isaacs} numerically on a multi-dimensional grid,
which we refer to as the ``oracle'' in this section. \new{The value function is represented as a grid with about $91$ million scalar values, taking 0.7 GB.}

\paragraph{Safety Filters}
We implement our robust rollout safety filter with a modified zonotope-based FRS scheme~\citep{bak2020zonoreach}\footnote{We use zonotopes since the computation of their Minkowski sum is light. We refer readers to~\citep{althoff2021set} for other representations for FRSs.}.
We also implement a direct rollout safety filter that checks gameplay trajectories:
$\shieldNaive(\state_\tdisc, \policyTask, \policyDstb, \khorizon) := \mathbbm{1} \{ \stateimag{\tdiscaux}{\tdisc} \notin \failure, \forall \tdiscaux \in \{1, \cdots, \khorizon+1\} \}$, where the zero disturbance in~\eqref{eq:imag_rollout} is replaced by disturbances from ISAACS $\policyDstb=\actorDstb$ or the oracle $\policyDstb=\policy^{\dstb*}$.
The safety filter is constructed by replacing $\valFunc(\state) \geq \epsilon$ with $\shieldNaive(\state, \policyTask, \policyDstb, \khorizon)=1$ and $\policy^{\ctrl*}(\state)$ with $\actorCtrl(\state)$ in~\eqref{eq:shield_original}.

\paragraph{Evaluation}
We evaluate 400 rollouts with initial states sampled uniformly from the state space
${\xSet = \{ \state \in \reals^5 \mid  p_x \!\in\! [0, 20],\; p_y \!\in\! [-0.6, 0.6],\; v \!\in\! [0.4, 2.0], \;\psi \!\in\! [-0.5\pi, 0.5 \pi], \;\delta \!\in\! [-0.35, 0.35] \}}$.\\
We compute the \emph{safe rate} as the fraction of trajectories that avoid the failure set for the full horizon.
\new{To more closely benchmark the predictions and performance of ISAACS against the oracle solution, we evaluate the critic and roll out the policies from each of the $91$ million cells in the oracle's grid.
}%
To evaluate different safety filters,
we use iterative LQR~\citep{li2004ilqr} as the task policy,
\new{with a barrier penalty to discourage violations},
and use the oracle disturbance policy with bound $\dstb_{\max}=0.1$ to attack the controller.
\new{We select 395 initial states from which the trained ISAACS safety policy can maintain safety against oracle disturbance, but the task policy results in constraint violations.}
In addition to the safe rate, we also compute the \emph{filter frequency}, which is the fraction of time steps at which the safety filter was triggered, averaged across all trajectories.

\subsection{Results}
\paragraph{Offline Adversarial RL}
\begin{figure}[!t]
\floatconts
    {fig:isaacs}
    {\vspace{-10mm}\caption{
        \textit{Left:} Comparison of safety controllers' robustness to disturbances.
        As the disturbance bound increases, controllers trained without disturbance or with DR rapidly degrade. The ISAACS controller trained against the largest adversarial disturbance suffers the least safety degradation, nearing the optimal (oracle) policy.
        \textit{Right:} ``Confusion plots'' of values and rollout outcomes for 2-D slices of the state space, with $v = 1, \psi = 0, \delta=0.03$.
        \textit{Top:} learned safety critic can wrongly predict some rollout outcomes, leading to inaccuracies in the estimated safe set boundary.
        \textit{Middle:} learned ISAACS safety policy achieves near-optimal success but is occasionally suboptimal near the safe set boundary.
        \textit{Bottom:} direct policy rollout using the learned disturbance can lead to over-optimistic predictions.
        \vspace{-10mm}
    }}
    {
        \includegraphics[width=\textwidth]{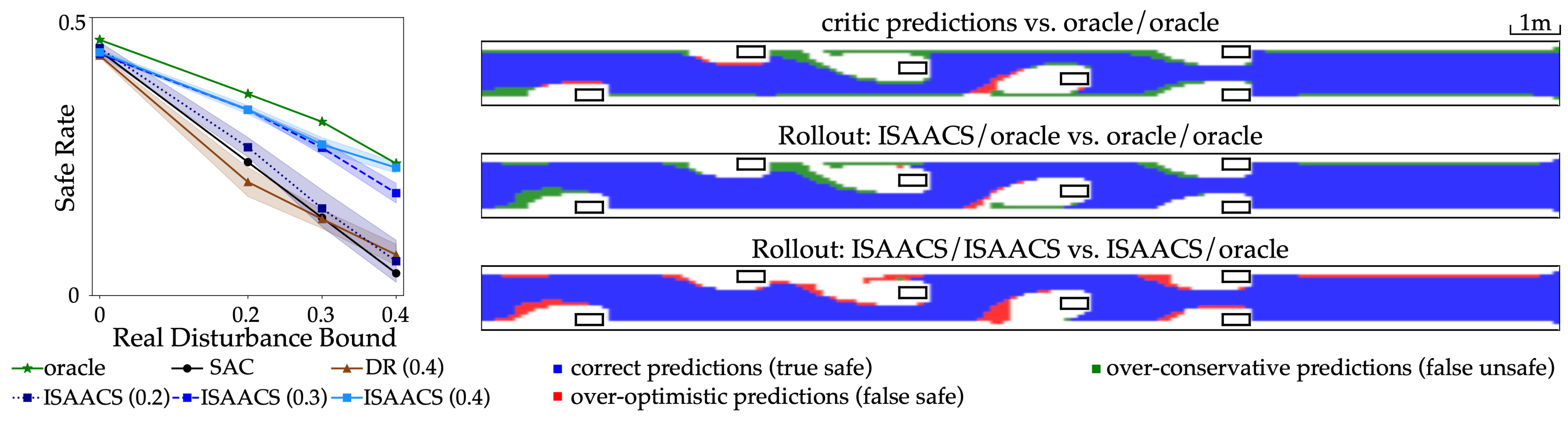}
    }
\end{figure}

We first evaluate the ISAACS controller's robustness. \figureref{fig:isaacs} (left) shows the \new{average} safe rate of ISAACS and other safety policies under different disturbance bounds \new{over three random seeds with the shaded region for one standard deviation}.
ISAACS' robustness improves as the bound used in the training increases, indicating that the learned disturbance policy approximates the worst-case uncertainty realization. 
Further, ISAACS outperforms single-agent RL (even with DR) by a large margin. 
DR optimizes against the average among all possible disturbances and is less robust
to worst-case realizations.
Finally, when trained with the highest disturbance bound, ISAACS presents comparable robustness to the oracle safety policy.

\figureref{fig:isaacs} (right) shows confusion matrix color plots of value and rollout predictions across 2-D slices of the state space when $v = 1, \psi = 0, \delta=0.03$.
\new{The ISAACS critic (top plot) achieves 1.4\% false-safe rate (red region),
which is remarkable given that it uses over $1,000\times$ fewer parameters than the numerical HJI oracle.
This contrasts with a more conservative 20.4\% false-unsafe rate (green region).
When pitted against the oracle worst-case disturbance, the learned ISAACS controller (middle plot)
loses safety from 10.4\% of true safe states, where the oracle controller succeeds.
If we replace the oracle disturbance with the learned one (bottom plot), 12\% of states where the ISAACS controller fails in the true worst case are mispredicted as safe by the ISAACS gameplay rollout. This fallibility motivates the use of a \emph{robust} rollout of the learned ISAACS controller under all $\dstb\in\dSet$ rather than relying on a direct rollout of the ISAACS controller and disturbance.
}

\paragraph{Online Robust Rollout Safety Filter}
\begin{wrapfigure}[12]{r}{0.6\textwidth}
    \center
    \vspace{-12mm}
    \includegraphics[width=.6\textwidth]{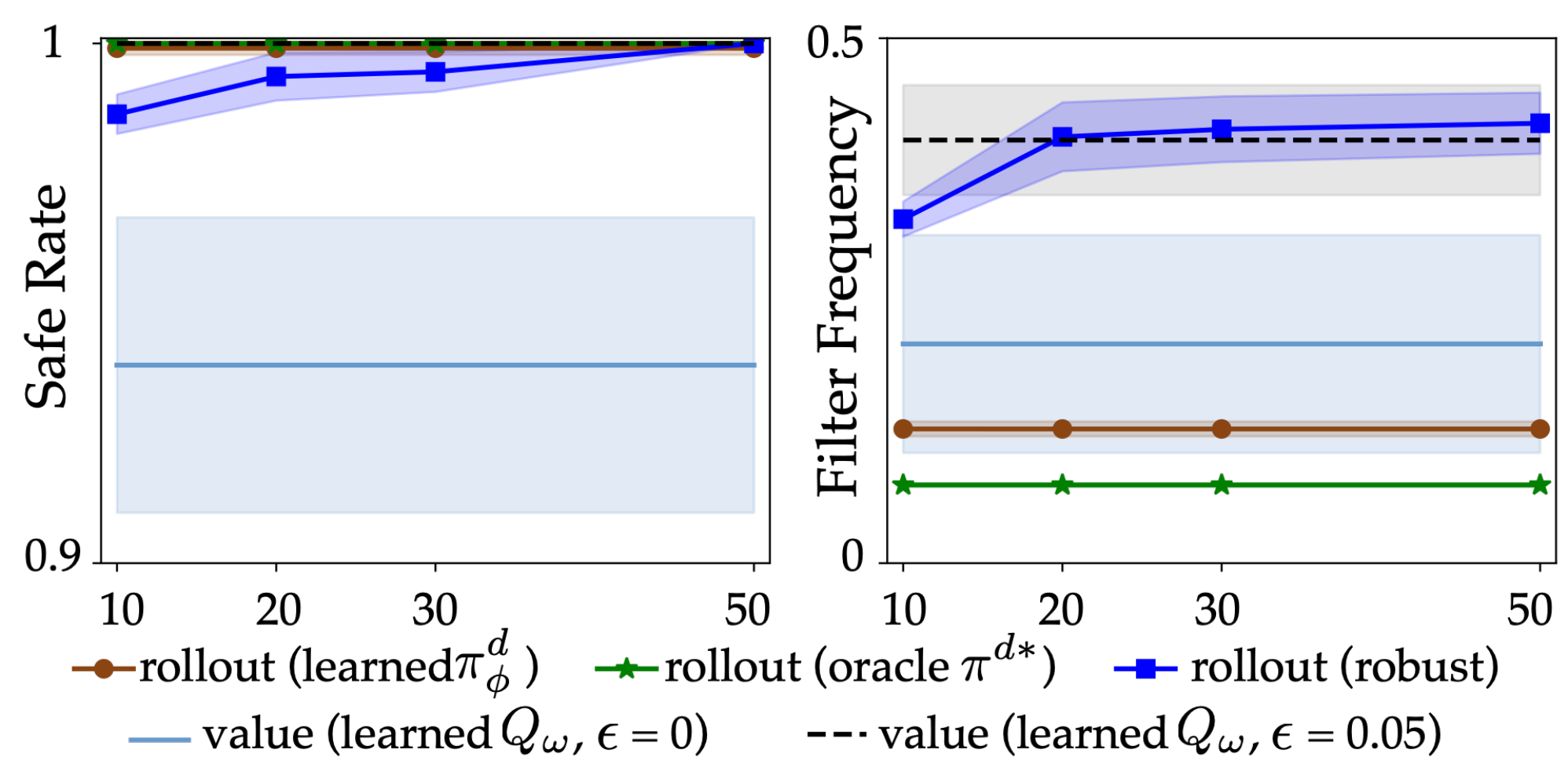}
    \vspace{-10mm}
    \caption{
        Safe rate and conservativeness of different safety filters. 
        A robust rollout-based safety filter with horizon $\khorizon=50$ achieves zero-violation safety.
    }
    \label{fig:shield}
\end{wrapfigure}
\looseness=-1
\new{To account for linearization error, we expand the disturbance bound in each dimension by 5\% for the purposes of computing the zonotopic forward-reachable set.
\figureref{fig:shield} shows the average safe rate and filter frequency over three random seeds with shaded regions for one standard deviation.}
Our proposed robust rollout-based safety filter \new{achieves a perfect 100\% safety rate for a long enough lookahead horizon (50 time steps), in contrast with the 99\% safe rate obtained through a direct gameplay rollout safety filter
and the 94\% obtained by naively filtering with the sign of the learned safety critic.
The value-based safety filter can be improved by introducing a small threshold $\epsilon$ = 0.05 to mitigate approximation errors, even achieving perfect (empirical) safety.
Yet, the value-based filter lacks theoretical guarantees, and its threshold needs manual tuning, difficult before real deployment.
Despite its lack of guarantees, the direct gameplay rollout filter performs remarkably well, approaching the oracle in both safe rate and filter frequency.
}%
\new{
The robust rollout-based safety filter has a more conservative activation frequency, similar
to the one achieved by 
the manually thresholded value-based safety filter.
Importantly, the robust rollout filter provides strong, clear-cut guarantees, which are of practical importance for the safe deployment of autonomous systems.
}

\section{Conclusion}
\looseness=-1
We propose ISAACS, an adversarial reinforcement learning method for offline safety policy synthesis, whose learned policies can be used to build robust safety-certified control strategies at runtime.
We prove safety guarantees for an ISAACS-based safety filter using robust policy rollouts under bounded uncertainty.
We demonstrate experimentally that our proposed offline training has comparable performance to numerical methods in a 5-D race car simulator,
and the runtime safety filter achieves a perfect zero-violation safe rate with moderate added conservativeness.
These results open a promising research avenue for scalable synthesis of robust safety strategies using neural networks.

\acks{
This work is supported in part by the Google Research Scholar Award.
The authors thank Jie Tan and Wenhao Yu for their thoughtful suggestions and valuable insights.
}
\bibliography{ref}

\begin{thebibliography}{34}
\providecommand{\natexlab}[1]{#1}
\providecommand{\url}[1]{\texttt{#1}}
\expandafter\ifx\csname urlstyle\endcsname\relax
  \providecommand{\doi}[1]{doi: #1}\else
  \providecommand{\doi}{doi: \begingroup \urlstyle{rm}\Url}\fi

\bibitem[Alpern and Schneider(1985)]{alpern1985defining}
Bowen Alpern and Fred~B. Schneider.
\newblock Defining liveness.
\newblock 21\penalty0 (4):\penalty0 181--185, 1985.
\newblock ISSN 00200190.
\newblock \doi{10.1016/0020-0190(85)90056-0}.
\newblock URL
  \url{https://linkinghub.elsevier.com/retrieve/pii/0020019085900560}.

\bibitem[Althoff et~al.(2021)Althoff, Frehse, and Girard]{althoff2021set}
Matthias Althoff, Goran Frehse, and Antoine Girard.
\newblock Set propagation techniques for reachability analysis.
\newblock \emph{Annual Review of Control, Robotics, and Autonomous Systems},
  4\penalty0 (1):\penalty0 369--395, 2021.
\newblock \doi{10.1146/annurev-control-071420-081941}.

\bibitem[Ames et~al.(2014)Ames, Grizzle, and Tabuada]{ames2014control}
Aaron~D Ames, Jessy~W Grizzle, and Paulo Tabuada.
\newblock {Control Barrier Function Based Quadratic Programs with Application
  to Adaptive Cruise Control}.
\newblock In \emph{{53rd IEEE Conference on Decision and Control}}, pages
  6271--6278. IEEE, 2014.

\bibitem[Ames et~al.(2019)Ames, Coogan, Egerstedt, Notomista, Sreenath, and
  Tabuada]{ames2019control}
Aaron~D. Ames, Samuel Coogan, Magnus Egerstedt, Gennaro Notomista, Koushil
  Sreenath, and Paulo Tabuada.
\newblock Control barrier functions: Theory and applications.
\newblock In \emph{Proceedings of the 18th European Control Conference (ECC)},
  pages 3420--3431, 2019.
\newblock \doi{10.23919/ECC.2019.8796030}.

\bibitem[Bak(2020)]{bak2020zonoreach}
Stanley Bak.
\newblock Quick zono reach, 2020.
\newblock URL \url{https://github.com/stanleybak/quickzonoreach}.

\bibitem[Bansal and Tomlin(2021)]{bansal2021deepreach}
Somil Bansal and Claire~J. Tomlin.
\newblock Deepreach: A deep learning approach to high-dimensional reachability.
\newblock In \emph{Proceedings of the IEEE International Conference on Robotics
  and Automation (ICRA)}, pages 1817--1824, 2021.
\newblock \doi{10.1109/ICRA48506.2021.9561949}.

\bibitem[Bansal et~al.(2017)Bansal, Chen, Herbert, and
  Tomlin]{bansal2017hamilton}
Somil Bansal, Mo~Chen, Sylvia Herbert, and Claire~J. Tomlin.
\newblock Hamilton-jacobi reachability: A brief overview and recent advances.
\newblock In \emph{Proceedings of the IEEE Annual Conference on Decision and
  Control (CDC)}, pages 2242--2253, 2017.
\newblock \doi{10.1109/CDC.2017.8263977}.

\bibitem[Bastani and Li(2021)]{bastani2021safe}
Osbert Bastani and Shuo Li.
\newblock Safe reinforcement learning via statistical model predictive
  shielding.
\newblock In \emph{Proceedings of Robotics: Science and Systems}, Virtual, 7
  2021.
\newblock \doi{10.15607/RSS.2021.XVII.026}.

\bibitem[Bharadhwaj et~al.(2021)Bharadhwaj, Kumar, Rhinehart, Levine, Shkurti,
  and Garg]{bharadhwaj2021conservative}
Homanga Bharadhwaj, Aviral Kumar, Nicholas Rhinehart, Sergey Levine, Florian
  Shkurti, and Animesh Garg.
\newblock Conservative safety critics for exploration.
\newblock In \emph{Proceedings of the 9th International Conference on Learning
  Representations, {ICLR}, Virtual Event, Austria, May 3-7, 2021}, 2021.
\newblock URL \url{https://openreview.net/forum?id=iaO86DUuKi}.

\bibitem[Bradbury et~al.(2018)Bradbury, Frostig, Hawkins, Johnson, Leary,
  Maclaurin, Necula, Paszke, Vander{P}las, Wanderman-{M}ilne, and
  Zhang]{bradbury2018jax}
James Bradbury, Roy Frostig, Peter Hawkins, Matthew~James Johnson, Chris Leary,
  Dougal Maclaurin, George Necula, Adam Paszke, Jake Vander{P}las, Skye
  Wanderman-{M}ilne, and Qiao Zhang.
\newblock {JAX}: composable transformations of {P}ython+{N}um{P}y programs,
  2018.
\newblock URL \url{http://github.com/google/jax}.

\bibitem[Bui et~al.(2022)Bui, Giovanis, Chen, and
  Shriraman]{bui2022optimizeddp}
Minh Bui, George Giovanis, Mo~Chen, and Arrvindh Shriraman.
\newblock {OptimizedDP}: An efficient, user-friendly library for optimal
  control and dynamic programming, 2022.
\newblock URL \url{https://arxiv.org/abs/2204.05520}.

\bibitem[Choi et~al.(2021)Choi, Lee, Sreenath, Tomlin, and
  Herbert]{choi2021robust}
Jason~J. Choi, Donggun Lee, Koushil Sreenath, Claire~J. Tomlin, and Sylvia~L.
  Herbert.
\newblock Robust control barrier-value functions for safety-critical control.
\newblock In \emph{Proceedings of the 60th IEEE Conference on Decision and
  Control (CDC)}, pages 6814--6821, 2021.
\newblock \doi{10.1109/CDC45484.2021.9683085}.

\bibitem[{Fisac} et~al.(2019){Fisac}, {Akametalu}, {Zeilinger}, {Kaynama},
  {Gillula}, and {Tomlin}]{fisac2019AGS}
Jaime~F. {Fisac}, Anayo~K. {Akametalu}, Melanie~N. {Zeilinger}, Shahab
  {Kaynama}, Jeremy {Gillula}, and Claire~J. {Tomlin}.
\newblock A general safety framework for learning-based control in uncertain
  robotic systems.
\newblock \emph{IEEE Transactions on Automatic Control}, 64\penalty0
  (7):\penalty0 2737--2752, 2019.
\newblock \doi{10.1109/TAC.2018.2876389}.

\bibitem[Fisac et~al.(2019)Fisac, Lugovoy, Rubies-Royo, Ghosh, and
  Tomlin]{fisac2019bridging}
Jaime~F. Fisac, Neil~F. Lugovoy, Vicenç Rubies-Royo, Shromona Ghosh, and
  Claire~J. Tomlin.
\newblock Bridging hamilton-jacobi safety analysis and reinforcement learning.
\newblock In \emph{Proceedings of the International Conference on Robotics and
  Automation (ICRA)}, pages 8550--8556, 2019.
\newblock \doi{10.1109/ICRA.2019.8794107}.

\bibitem[Haarnoja et~al.(2018)Haarnoja, Zhou, Abbeel, and
  Levine]{haarnoja2018sac}
Tuomas Haarnoja, Aurick Zhou, Pieter Abbeel, and Sergey Levine.
\newblock Soft actor-critic: Off-policy maximum entropy deep reinforcement
  learning with a stochastic actor.
\newblock In \emph{Proceedings of the 35th International Conference on Machine
  Learning}, volume~80 of \emph{Proceedings of Machine Learning Research},
  pages 1861--1870. PMLR, 7 2018.
\newblock URL \url{https://proceedings.mlr.press/v80/haarnoja18b.html}.

\bibitem[Hsu et~al.(2021)Hsu, Rubies-Royo, Tomlin, and Fisac]{hsu2021safety}
Kai-Chieh Hsu, Vicenç Rubies-Royo, Claire~J. Tomlin, and Jaime~F. Fisac.
\newblock Safety and liveness guarantees through reach-avoid reinforcement
  learning.
\newblock In \emph{Proceedings of Robotics: Science and Systems}, Virtual, 7
  2021.
\newblock \doi{10.15607/RSS.2021.XVII.077}.

\bibitem[Hsu et~al.(2023)Hsu, Ren, Nguyen, Majumdar, and
  Fisac]{hsuzen2022sim2lab2real}
Kai-Chieh Hsu, Allen~Z. Ren, Duy~P. Nguyen, Anirudha Majumdar, and Jaime~F.
  Fisac.
\newblock Sim-to-lab-to-real: Safe reinforcement learning with shielding and
  generalization guarantees.
\newblock \emph{Artificial Intelligence}, 314:\penalty0 103811, 2023.
\newblock ISSN 0004-3702.
\newblock \doi{https://doi.org/10.1016/j.artint.2022.103811}.
\newblock URL
  \url{https://www.sciencedirect.com/science/article/pii/S0004370222001515}.

\bibitem[Isaacs(1954)]{isaacs1954differential}
Rufus Isaacs.
\newblock Differential {{Games I}}: {{Introduction}}.
\newblock 1954.
\newblock URL \url{https://www.rand.org/pubs/research_memoranda/RM1391.html}.

\bibitem[Langson et~al.(2004)Langson, Chryssochoos, Rakovi{\'c}, and
  Mayne]{langson2004robust}
W.~Langson, I.~Chryssochoos, S.~V. Rakovi{\'c}, and D.~Q. Mayne.
\newblock Robust model predictive control using tubes.
\newblock \emph{Automatica}, 40\penalty0 (1):\penalty0 125--133, 2004.
\newblock ISSN 0005-1098.
\newblock \doi{10.1016/j.automatica.2003.08.009}.
\newblock URL
  \url{http://www.sciencedirect.com/science/article/pii/S0005109803002838}.

\bibitem[Li and Todorov(2004)]{li2004ilqr}
Weiwei Li and Emanuel Todorov.
\newblock Iterative linear quadratic regulator design for nonlinear biological
  movement systems.
\newblock In \emph{ICINCO (1)}, pages 222--229, 2004.

\bibitem[Mattila et~al.(2015)Mattila, Mo, and Murray]{mattila2015iterative}
Robert Mattila, Yilin Mo, and Richard~M. Murray.
\newblock An iterative abstraction algorithm for reactive
  correct-by-construction controller synthesis.
\newblock In \emph{2015 54th {{IEEE Conference}} on {{Decision}} and
  {{Control}} ({{CDC}})}, pages 6147--6152, 2015.
\newblock \doi{10.1109/CDC.2015.7403186}.

\bibitem[Mehta et~al.(2020)Mehta, Diaz, Golemo, Pal, and
  Paull]{mehta2020active}
Bhairav Mehta, Manfred Diaz, Florian Golemo, Christopher~J. Pal, and Liam
  Paull.
\newblock Active domain randomization.
\newblock In \emph{Proceedings of the Conference on Robot Learning}, volume
  100, pages 1162--1176, 30 Oct--01 Nov 2020.
\newblock URL \url{https://proceedings.mlr.press/v100/mehta20a.html}.

\bibitem[Mitchell et~al.(2005)Mitchell, Bayen, and
  Tomlin]{mitchell2005timedependent}
Ian~M. Mitchell, Alexandre~M. Bayen, and Claire~J. Tomlin.
\newblock A time-dependent {{Hamilton-Jacobi}} formulation of reachable sets
  for continuous dynamic games.
\newblock \emph{IEEE Transactions on Automatic Control}, 50\penalty0
  (7):\penalty0 947--957, 2005.
\newblock ISSN 1558-2523.
\newblock \doi{10.1109/TAC.2005.851439}.

\bibitem[Nguyen and Sreenath(2016)]{nguyen2016optimal}
Quan Nguyen and Koushil Sreenath.
\newblock Optimal robust time-varying safety-critical control with application
  to dynamic walking on moving stepping stones.
\newblock In \emph{Proceedings of the {ASME} 2016 {Dynamic Systems and Control
  Conference}.}, page V002T28A005. {American Society of Mechanical Engineers},
  2016.
\newblock ISBN 978-0-7918-5070-1.
\newblock \doi{10.1115/DSCC2016-9910}.
\newblock URL
  \url{https://asmedigitalcollection.asme.org/DSCC/proceedings/DSCC2016/50701/Minneapolis,%20Minnesota,%20USA/231011}.

\bibitem[Pinto et~al.(2017)Pinto, Davidson, Sukthankar, and
  Gupta]{pinto2017robust}
Lerrel Pinto, James Davidson, Rahul Sukthankar, and Abhinav Gupta.
\newblock Robust adversarial reinforcement learning.
\newblock In \emph{Proceedings of the 34th International Conference on Machine
  Learning - Volume 70}, pages 2817--2826, 2017.

\bibitem[Silver et~al.(2017)Silver, Hubert, Schrittwieser, Antonoglou, Lai,
  Guez, Lanctot, Sifre, Kumaran, Graepel, Lillicrap, Simonyan, and
  Hassabis]{silver2017alphazero}
David Silver, Thomas Hubert, Julian Schrittwieser, Ioannis Antonoglou, Matthew
  Lai, Arthur Guez, Marc Lanctot, Laurent Sifre, Dharshan Kumaran, Thore
  Graepel, Timothy Lillicrap, Karen Simonyan, and Demis Hassabis.
\newblock Mastering chess and shogi by self-play with a general reinforcement
  learning algorithm, 2017.

\bibitem[Squires et~al.(2018)Squires, Pierpaoli, and
  Egerstedt]{squires2018constructive}
Eric Squires, Pietro Pierpaoli, and Magnus Egerstedt.
\newblock Constructive barrier certificates with applications to fixed-wing
  aircraft collision avoidance.
\newblock In \emph{2018 {{IEEE Conference}} on {{Control Technology}} and
  {{Applications}} ({{CCTA}})}, pages 1656--1661. {IEEE}, 2018.
\newblock ISBN 978-1-5386-7698-1.
\newblock \doi{10.1109/CCTA.2018.8511342}.
\newblock URL \url{https://ieeexplore.ieee.org/document/8511342/}.

\bibitem[Thananjeyan et~al.(2021)Thananjeyan, Balakrishna, Nair, Luo,
  Srinivasan, Hwang, Gonzalez, Ibarz, Finn, and
  Goldberg]{thananjeyan2021recovery}
Brijen Thananjeyan, Ashwin Balakrishna, Suraj Nair, Michael Luo, Krishnan
  Srinivasan, Minho Hwang, Joseph~E. Gonzalez, Julian Ibarz, Chelsea Finn, and
  Ken Goldberg.
\newblock Recovery rl: Safe reinforcement learning with learned recovery zones.
\newblock \emph{IEEE Robotics and Automation Letters}, 6\penalty0 (3):\penalty0
  4915--4922, 2021.
\newblock \doi{10.1109/LRA.2021.3070252}.

\bibitem[Tobin et~al.(2017)Tobin, Fong, Ray, Schneider, Zaremba, and
  Abbeel]{tobin2017domain}
Josh Tobin, Rachel Fong, Alex Ray, Jonas Schneider, Wojciech Zaremba, and
  Pieter Abbeel.
\newblock Domain randomization for transferring deep neural networks from
  simulation to the real world.
\newblock In \emph{Proceedings of the IEEE/RSJ International Conference on
  Intelligent Robots and Systems (IROS)}, pages 23--30, 2017.
\newblock \doi{10.1109/IROS.2017.8202133}.

\bibitem[Vinitsky et~al.(2020)Vinitsky, Du, Parvate, Jang, Abbeel, and
  Bayen]{vinitsky2020robust}
Eugene Vinitsky, Yuqing Du, Kanaad Parvate, Kathy Jang, Pieter Abbeel, and
  Alexandre Bayen.
\newblock Robust reinforcement learning using adversarial populations, 2020.

\bibitem[Wabersich and Zeilinger(2018)]{wabersich2018linear}
Kim~P. Wabersich and Melanie~N. Zeilinger.
\newblock Linear model predictive safety certification for learning-based
  control.
\newblock In \emph{2018 IEEE Conference on Decision and Control (CDC)}, pages
  7130--7135. IEEE, 2018.

\bibitem[Xu et~al.(2015)Xu, Tabuada, Grizzle, and Ames]{xu2015robustness}
Xiangru Xu, Paulo Tabuada, Jessy~W. Grizzle, and Aaron~D. Ames.
\newblock Robustness of control barrier functions for safety critical control.
\newblock 48\penalty0 (27):\penalty0 54--61, 2015.
\newblock ISSN 24058963.
\newblock \doi{10.1016/j.ifacol.2015.11.152}.
\newblock URL
  \url{https://linkinghub.elsevier.com/retrieve/pii/S2405896315024106}.

\bibitem[Zeng et~al.(2021)Zeng, Zhang, Li, and
  Sreenath]{zeng2021safetycritical}
Jun Zeng, Bike Zhang, Zhongyu Li, and Koushil Sreenath.
\newblock Safety-critical control using optimal-decay control barrier function
  with guaranteed point-wise feasibility.
\newblock In \emph{Proceedings of the American Control Conference (ACC)}, pages
  3856--3863, 2021.
\newblock \doi{10.23919/ACC50511.2021.9482626}.

\bibitem[Zrnic et~al.(2021)Zrnic, Mazumdar, Sastry, and Jordan]{tijana2021who}
Tijana Zrnic, Eric Mazumdar, Shankar Sastry, and Michael Jordan.
\newblock Who leads and who follows in strategic classification?
\newblock In \emph{Advances in Neural Information Processing Systems},
  volume~34, pages 15257--15269, 2021.
\newblock URL
  \url{https://proceedings.neurips.cc/paper/2021/file/812214fb8e7066bfa6e32c626c2c688b-Paper.pdf}.

\end{thebibliography}
\end{document}